\documentclass[onecolumn]{IEEEtran}
\usepackage{acro}
\usepackage[utf8]{inputenc}
\usepackage[ruled,vlined,linesnumbered]{algorithm2e}
\usepackage{derivative}
\usepackage{amsfonts, amsmath, amssymb}
\usepackage{array}
\usepackage{bm}
\usepackage{cite}
\usepackage{color}
\usepackage{float}
\usepackage{subfig}
\usepackage{hyperref}
\usepackage{url}
\usepackage{multirow}
\usepackage{booktabs}
\usepackage{graphicx}
\graphicspath{./figures/}
\usepackage[switch]{lineno}

\newcommand{\fig}[1]{Fig.~\ref{#1}}
\newcommand{\tab}[1]{Tab.~\ref{#1}}
\newcommand{\sect}[1]{Sec.~\ref{#1}}

\newcommand{\ve}[1]{\boldsymbol{\mathbf{#1}}}
\newcommand{\R}{\mathbb{R}}
\newcommand{\Sx}{\mathcal{S}}
\newcommand{\K}{\mathcal{K}}
\newcommand{\Jx}{\mathbf{J}}
\newcommand{\Jcal}{\mathcal{J}}

\DeclareMathOperator*{\argmax}{arg\,max}

\title{One-shot Learning for Autonomous Aerial Manipulation}

\author{
    Claudio Zito$^{1*}$, and Eliseo Ferrante$^{1}$
    \\
    $^{1}$ Technology Innovation Institute (TII), Abu Dhabi, United Arab Emirates\\
    Claudio.Zito\!\atsign tii.ae\\
    $^*$ Correspondent author
}

\newcommand\atsign{@}

\begin{document}
    \maketitle

\begin{abstract}
    This paper is concerned with learning transferable contact models for aerial manipulation tasks. We investigate a contact-based approach for enabling unmanned aerial vehicles with cable-suspended passive grippers to compute the attach points on novel payloads for aerial transportation. This is the first time that the problem of autonomously generating contact points for such tasks has been investigated. Our approach builds on the underpinning idea that we can learn a probability density of contacts over objects' surfaces from a single demonstration. We enhance this formulation for encoding aerial transportation tasks while maintaining the one-shot learning paradigm without handcrafting task-dependent features or employing ad-hoc heuristics; the only prior is extrapolated directly from a single demonstration. Our models only rely on the geometrical properties of the payloads computed from a point cloud, and they are robust to partial views. The effectiveness of our approach is evaluated in simulation, in which one or three quadropters are requested to transport previously unseen payloads along a desired trajectory. The contact points and the quadroptors configurations are computed on-the-fly for each test by our apporach and compared with a baseline method, a modified grasp learning algorithm from the literature. Empirical experiments show that the contacts generated by our approach yield a better controllability of the payload for a transportation task. We conclude this paper with a discussion on the strengths and limitations of the presented idea, and our suggested future research directions.   
\end{abstract}

\section{Introduction}

%In the words of the fictional character Dr Robert Ford "a simple handshake would give them away". As often happens, science fiction takes inspiration from reality. In this particular case, both fiction and reality agree on identifying one of the most challenging problems in robotics: how to manipulate objects. 
There is evidence that humans possess an internal model of physical interactions that enables us to grasp, lift, pull, or push objects of diverse nature in various contexts~\cite{doi:10.1152/jn.2002.88.2.942}. It is also evident that such internal models are constructed over time as an accumulation of experience as opposed to an inherent comprehension of physics.
In this paper, we investigate an internal model for enabling autonomous quadrotors equipped with cable-suspended passive grippers to generate contacts with unknown payloads for aerial transportation. By unknown payloads, we mean that our approach does not require a full CAD model of the object or information relative to its physical properties, such as its centre of mass (CoM) or friction coefficients. We only rely on geometric features extrapolated from vision. As the human's internal model, our solution is not failure-free but provides a way to generate candidate grasps even when no information is available. 

The hype for aerial manipulation has reached a high-fever pitch. Unmanned aerial vehicles (UAVs), such as quadrotors, have been recently at the centre of attention of the scientific community as the next means of autonomy. Their dynamic simplicity, manoeuvrability and high performance make them ideal for many applications ranging from surveillance to emergency response. More recently, such systems have been investigated for aerial transportation of payloads by towed cables. Small-size single and multiple quadrotors have been employed for load transportation and deployment by designing control laws for minimum swing and oscillation of the payload \cite{loianno_cooperative_2021,loianno_pcmpc_2021,kumar_diff_flatness_2013}. 
Although in their dawn, current approaches disregard the generation of the contact points. Single drones assume point-mass loads to simplify the effects of the dynamics, and multiple drones are manually attached nearby the vertices or edges of the load, following the intuition of maximising the moments exercised on the object. No aerial system is capable of generating on-the-fly contacts, and it is not clear how the proposed controllers can cope with different payloads or contact configurations.  

On the other hand, we have witnessed a growing interest in robot grasping and manipulation tasks in the last decade \cite{6672028,10.3389/frobt.2020.00008}. Although we are still far from robots freely manipulating arbitrary objects, several promising solutions have been proposed over the years \cite{9619964,brahmbhatt2019contactgrasp}. Google employed a dozen robots interacting in parallel with their own environment to learn how to predict what happens when they move objects around \cite{google_research}. However, collecting such a large amount of data for any task is very hard, and many researchers have focussed on more practical solutions. Additionally, the task of aerial manipulation presents another significant challenge--failed attempts may irreparably damage the payload and the AUVs. A more appealing approach is to learn models in a one-shot or a few-shot fashion when possible. Such approaches typically employ generative models which learn probability densities from demonstrations. By providing one or a few examples of a good solution, we substantially reduce the searching effort when facing novel contexts. Furthermore, when the learning space is constructed over local features, the models tend to have a good generalisation capability within and across object categories \cite{kopicki2016}.  
 
In \cite{kopicki2016,arruda_generative_2019},  the authors formulate dexterous grasps for a humanoid robot as contacts between the robot's manipulative links and the object's surface. Only local surface properties of the object's geometrics are required to learn the model, such as curvatures on the (local) contact patches. The model learns grasp types but does not encode task-dependent features or physical properties of the objects, and it has only been demonstrated for pick-and-place tasks. For different tasks beyond pick-and-place, the authors enable the possibility of encoding handcrafted loss functions, which requires (a rarely available) insight knowledge by the user on the specific task. In \cite{zito_push_2018, zito_push_2021}, this approach has been extended to enable a robot to make predictions over push operations for novel objects. One-shot learning is utilised for identifying the contacts between the pusher and the object, and between the object and the environment, but an extra \emph{motion model} needs to be learned from real or simulated experience in order to make the predictions. In practice, the motion models encode the task. They are designed as conditional probabilities given the initial contact models, following the general idea that making predictions on familiar initial conditions will yield more robust solutions. A few dozens of examples are sufficient to learn a motion model for planar push operations since the environment constrains the motion. Such luxury is not available for aerial manipulation, and it would require us to approximate the drone-payload system dynamics to make any prediction.

\begin{figure}[t]
    \centering{
    \includegraphics[width=.32\textwidth]{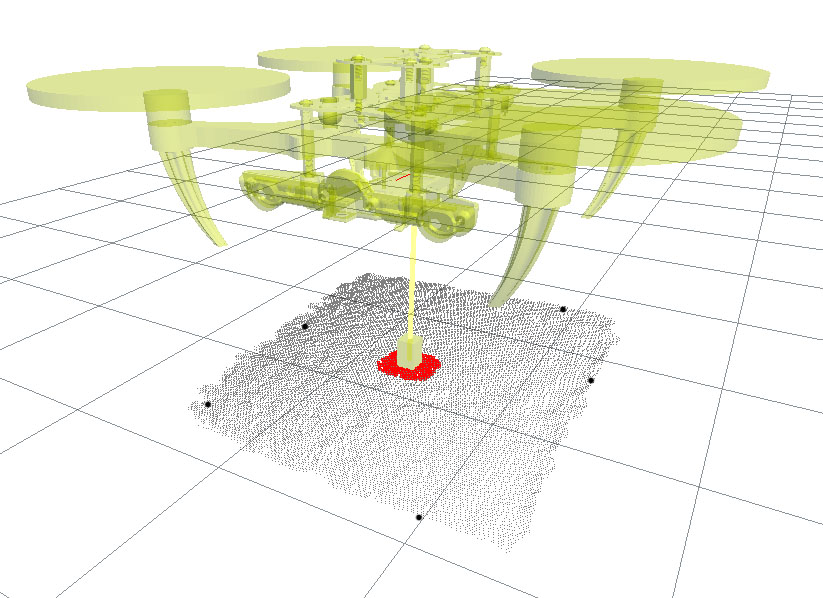}
    \includegraphics[width=.25\textwidth]{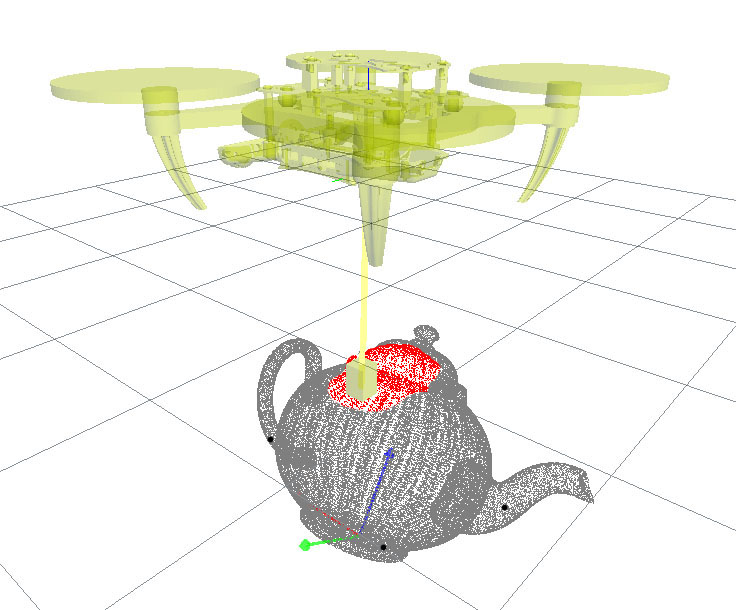}
    \includegraphics[width=.32\textwidth]{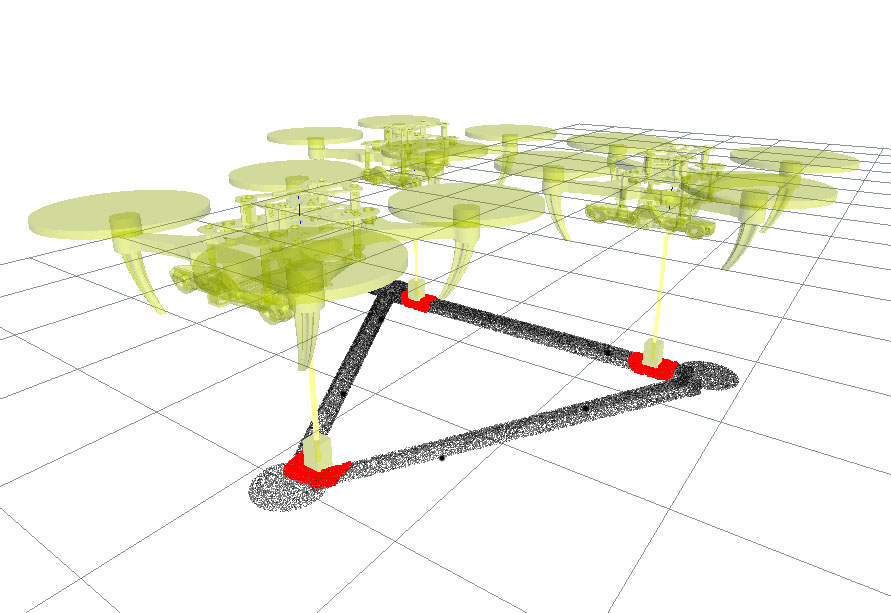}
    }
    \caption{Learned contact models. (left) shows a central contact with a flat surface, i.e. top face of the Fedex box, (middle) shows a curved contact on a teapot, and (right) a triangular configuration for three drones. The red points represent the sub-sampled local area of the surface used to learn the contact model. Black points represent the five highest of the local features used to learn the task. The global reference frame of the object, e.g. at the base of the teapot in (b), is not considered in the learning.}
    \label{fig:contact_models}
\end{figure}

In contrast, this paper proposes one-shot learning for aerial manipulation in which contacts are not solely learned from local features but also encode intrinsic task-dependent knowledge without the need for handcrafted features, as visible in \fig{fig:contact_models}. The underpinning idea is simple. Imagine having to connect a load to a single drone. If the load is a box with uniform mass distribution, probably the best thing we could do is attach the drone in the middle of the top facing surface, just above the payload's CoM. In order to find the desired spot, we do not need to have a perfect model of the object or see it in its completeness. A view on its entireness of the top facing surface would be sufficient for estimating its centre.
Nevertheless, counterexamples come to mind where merely teaching the robot to attach the payload above its CoM may not be a robust strategy; a package with overlapping ridges makes the desired contact area irregular and unsuitable for reliable contact. If we have taught the robot to connect over flat surfaces, we desire to maintain this property on novel objects. Hence, our approach weights the contact's local shape versus its relative location over the visible payload's surface. The latter is implemented as a probability distribution over the distances between the taught contact regions and sampled patches that describe the general shape of the visible geometry of the payload, e.g., flat surfaces, edges and corners. The main drawback of this approach is that we will need to learn a new contact model for each desired task and contact type. However, the framework is capable of generalising over different shaped payloads for the same task and contact type as presented in \sect{sec:learning}. 

We evaluate our proposed approach in a set of empirical experiments and compare the results against a baseline method. First, we have modified the approach in \cite{kopicki2016} to cope with a single or multiple UAVs equipped with cable-suspended passive grippers, in lieu of a dexterous manipulator. From the same training data we learned contact models for the baseline and our approach. Then, we use the learned models to generate candidate contact points for four test payloads. The contacts have been evaluated in a simulated environment for a transportation task, in which the UAVs need to lift the payload and transport it along a desired trajectory.

The rest of the paper is structured as follows. In \sect{sec:background}, we introduce the problem formulation in terms of object-centric representation of the contacts and the robot mechanics. Section~\ref{sec:learning} describes how we learn the models needed for aerial manipulation. Section~\ref{sec:transferring} presents how we infer contacts on novel shapes. The experimental evaluation is presented in \sect{sec:results} and our results are discussed in \sect{sec:conclusion}. We conclude with our final remarks about the strengths and limitations of this work and future research directions. 

\section{Problem formulation}\label{sec:background}

%In this section, we will introduce our formulation of the problem and building blocks used in our method described in~\sect{sec:learning}.

%\subsection{Notation}\label{sec:notation}

Let us begin from defining our notation. Vectors will be consider column vectors and written in bold letters. Matrix will be written as capital letters. We also denote by $SE(3)=\mathbb{R}^3\times SO(3)$ the standard Euclidian group in a three-dimensional space, and by $SO(3)=\{R\in\R^{3x3}|R^\top R=\mathbf{I}_{3x3}, \det[R]=1\}$ the special orthogonal group representing rotations in the three-dimensional space.
To describe rigid body transformations or poses, we will use the following format of denoting $v=(\ve{p},\ve{q})\in SE(3)$ where $\ve{p}\in\R^3$ is the translational component and $\ve{q}\in SO(3)$ is the quaternion describing the rotational component. Without losing generality, we abuse of the bold notation for the quaternion since they are implemented as a column vector in $\R^4$.

\vspace{0.5cm}
\subsection{Mechanics}\label{sec:mechanics}

We consider a set of $N_q\geq1$ quadrotors with cable-suspended magnetic grippers. The inertia frame $\mathbb{I}$ is defined by the unit vectors $\ve{e}_x=[1,0,0]^\top$, $\ve{e}_y=[0,1,0]^\top$, and $\ve{e}_z=[0,0,1]^\top\in\R^3$, and the third vector is aligned opposite to the direction of gravity. For each quadrotor, we define a body-fixed frame $\mathcal{B}_n=[\ve{b}_x, \ve{b}_y, \ve{b}_z]$ located at the centre of mass (CoM) of the quadrotor with its third axis aligned upward.

The pose $b_n=(\ve{p}_n,\ve{q}_n)\in SE(3)$ describes the location of the CoM of the $n$-th quadrotor, $\ve{p}_n\in\R^3$, and its rotation with respect to the inertia frame, $\ve{q}_n\in SO(3)$.
The mass and the inertia matrix for each quadrotor are denoted as $m_n\in\R$ and $\Jx_n\in\R^{3x3}$, respectively. We also denote the control input for the quadrotor as the pair $(f_n, \ve{M}_n)$, where $f_n\in\R$ is the total thrust and $\ve{M}_n\in\R^3$ the generated moment with respect to its body frame. With respect to the inertia frame, the quadrotor can generate a thrust $fR(\ve{q})\ve{e}_z\in\R^3$, where $R(\ve{q})\in SO(3)$ is the equivalent rotation matrix to the quaternion $\ve{q}$.

Each quadrotor is equipped with a cable of length $l_n$ and let $\xi_n\in\Sx$ be the unit-vector representing the direction of the $n$-th cable pointing outward from the quadrotor's CoM to the gripper. Let $L_n\in SE(3)$ be the pose for the end-effector's link for the $n$-th quadrotor with respect to the inertia frame. The contact points are defined with respect to the payload's reference frame, $z\in SE(3)$, where $m_z\in\R$ and $\Jx_z\in\R^{3x3}$ represent the mass and the inertia matrix, respectively.   

\vspace{0.5cm}
\subsection{Surface features}\label{sec:features}

\begin{figure}[t]
    \centering
    \includegraphics[width=.99\textwidth]{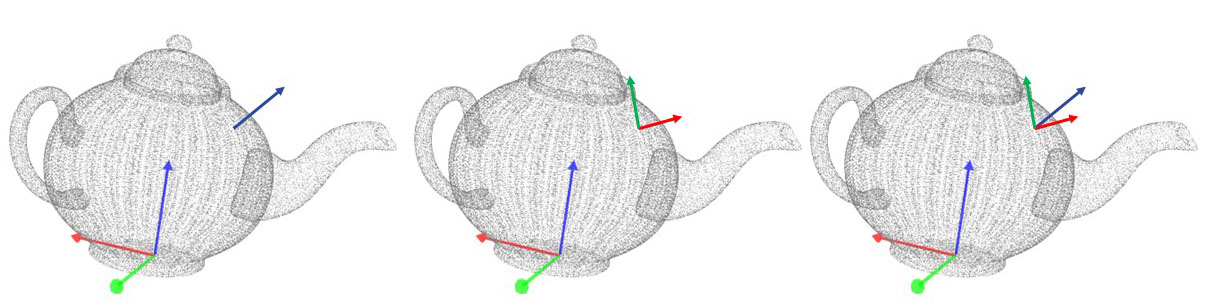}
    \caption{Computation of a surface feature on a teapot. (left) the full point cloud of the teapot with its reference frame. The normal of the sampled point is show as a blue arrow. (middle) the principal curvature directions are drown for the same point as red and green arrows. (right) the associated reference frame for the surface feature with respect to the object's reference frame.}
    \label{fig:surface_feature}
\end{figure}

Surface features encode the geometrical properties of an object and are derived from a 3D point cloud, as shown in \fig{fig:surface_feature}. We represent a surface feature as a pair $s=(v,\ve{r})\in SE(3) \times \R^2$, where $v=(\ve{p},\ve{q})\in SE(3)$ represents the pose of the surface feature $s$ and $\ve{r}\in\mathbb{R}^2$ is a vector of the surface descriptors. All poses denoted by $v$ are specified relative to the inertia frame $\mathcal{I}$.

To compute the pose $v\in SE(3)$, we first estimate the surface normal at point $\ve{p}$ using a PCA-based method \cite{kanatani2005geometry}. Surface descriptors correspond to the local principal curvatures around point $p$ \cite{spivak1999geometry}, which lie on the tangential plane to the object's surface and perpendicular to the surface normal at $p$. Let $k_1 \in \mathbb{R}^3$ be the direction of highest curvature, and $k_2 \in \mathbb{R}^3$ the direction of lowest curvature perpendicular to $k_1$. Let us also define $\ve{r} = [r_1, r_2]^\top \in \mathbb{R}^2$ as a 2D feature vector representing the curvatures along directions $k_1$ and $k_2$, respectively. The surface normal and principal directions form a body-fixed frame for the surface point $\ve{p}$ and enable us to compute the 3D orientation $\ve{q}$ that is associated to the point.

\vspace{0.5cm}
\subsection{Kernel density estimator} \label{sec:kde}

In this work, probability density functions are approximated via kernel density estimation (KDE) \cite{silverman1986density}, which are built around surface features (see~\sect{sec:features}). 
A kernel can be described by its mean point $\mu^s = (\mu_{p},\mu_{q},\mu_{r})$ and bandwidth $\sigma^s = (\sigma_{p},\sigma_{q},\sigma_{r})$:
\begin{equation} \label{eq:surface_feature_kernel}
\begin{aligned}
    \K(s\, &|\, \mu^s,\sigma^s)\ = \\ & \mathcal{N}_3(\ve{p}\, | \, \mu_{p},\sigma_{p})\ \Theta(\ve{q}\, |\, \mu_{q},\sigma_{q})\ \mathcal{N}_2(\ve{r}\, |\, \mu_{r},\sigma_{r})
\end{aligned}
\end{equation}
where $s=(v,\ve{r})=(\ve{p},\ve{q},\ve{r})$ is the surface feature being compared against the kernel, $\mathcal{N}_k$ is an $k$-variate Gaussian distribution, and $\Theta$ corresponds to a pair of antipodal von Mises-Fisher distributions forming a distribution similar to that of a Gaussian distribution for $SO(3)$ \cite{fisher1953dispersion}.

%In practice, each factor in $\K$ are computed as a truncated Gaussian approximation relative to the distance between two given features as follows:

%\begin{equation}
%    \mathcal{N}_3(\ve{p}|\mu_p,\!\sigma_p) \simeq
%    \begin{cases}
%    0,\!&\!\beta_p d_p(\ve{p},\mu_p,\!\sigma_p) \geq \delta_p\\
%    e^{-\beta_p d_p(\ve{p},\mu_p,\sigma_p)},\!&\!\beta_p d_p(\ve{p},\mu_p,\!\sigma_p) < \delta_p
%    \end{cases}
%\end{equation}
%\begin{equation}
%    \Theta(\ve{q}\,|\,\mu_q,\sigma_q) \simeq
%    \begin{cases}
%    0, & \beta_q d_q(\ve{q},\mu_q,\sigma_q) \geq \delta_q\\
%    e^{-\beta_q d_q(\ve{q},\mu_q,\sigma_q)}, & \beta_q %    \end{cases}
%\end{equation}
%\begin{equation}
%    \mathcal{N}_2(\ve{r}\,|\,\mu_r,\sigma_r) \simeq
%    \begin{cases}
%    0, & \beta_r d_r(\ve{r},\mu_r,\sigma_r) \geq \delta_r\\
%    e^{-\beta_r d_r(\ve{r},\mu_r,\sigma_r)}, & \beta_r d_r(\ve{r},\mu_r,\sigma_r) < \delta_r
%    \end{cases}
%\end{equation}

%where $d_p(\ve{p},\mu_{p},\sigma_{p}) = \frac{{|| \ve{p} - \mu_{p} ||}^2}{\sigma_{p}}$ is the Euclidiean distance between the surface feature point and the kernel point, $d_q(\ve{q},\mu_{q},\sigma_{q}) = \frac{1 - | \langle \ve{q}, \mu_{q} \rangle |}{\sigma_{q}}$ is the quaternion distance, $d_r(\ve{r},\mu_{r},\sigma_{r}) = (\ve{r} - \mu_{r})^\intercal D_{{\sigma_{r}}^{-1}} (\ve{r} - \mu_{r})$ is the principal curvature distance and $D_{{\sigma_{r}}^{-1}}$ is a diagonal matrix formed of the reciprocals of the surface descriptor bandwidths $\sigma_{r}$.

Given a set of $N_{s}$ surface features, the probability density in a region of space is computed as the local density of features in that region, as
\begin{equation} \label{eq:surface_feature_probability}
    P(\ve{p}, \ve{q}, \ve{r}) \equiv P(s) \simeq \sum_{i=1}^{N_{s}} w_i \K(s|s_i,\sigma^s)
\end{equation}
where $s_i$ corresponds to the $i$-{th} surface feature acting as a kernel, $w_i$ corresponds to its weighting with the constraint $\sum_{i=1}^{N_s} w_i\!=\!1$, and $\sigma^s$ is a user-defined bandwidth for the surface features.

\section{Learning contacts}\label{sec:learning}

We learn contacts from a single demonstration. We require a point cloud of the payload and the configuration of the quadrotors and their passive grippers. Partial views of the objects are sufficient for learning reliable models under the assumption that the surface in contact is fully visible. Figure~\ref{fig:contact_models} (left) shows the case in which only the top face of a box is visible at training time. Although, learning contacts is computational efficient as demonstrated in \cite{kopicki2016}, the time computation grows linearly with the number of links in contact, the number of triangles in the mesh representing the links, and the surface points considered. Furthermore, a new model needs to be learned for different drone configurations, contact types or tasks. 

\vspace{0.5cm}
\subsection{Object and task model}\label{sec:object_model}

The object model describes the composition of a point cloud in terms of its surface features distribution.
In the literature, this approach is used to learn the features distribution only nearby the contact area, e.g., \cite{kopicki2016,arruda_generative_2019,zito_push_2018}, while in \cite{zito_push_2021} the CoM of the object to be pushed is also estimated as a distribution from visible surface features. In contrast, we decouple the object model in two densities. The first describes the surface features distribution nearby the contacts, while the second encodes the location of these near-the-contact features with respect to other features in the visible point cloud, e.g., its corners and edges. 

At training time, we observe a set of contacts between the quadrotors' manipulative links and an object point cloud. Before learning a contact model (see~\sect{sec:contact_model}), we collect a set of $N_{O}$ features, $s_i=(v_i,\ve{r}_i)\in SE(3)\times\R^2$, within the surface in contact with the manipulative link. This enables us to compute a joint probability distribution as
\begin{equation} \label{eq:object_model}
    O(v, \ve{r}) \equiv P(v, \ve{r}) \simeq \sum_{i=1}^{N_{O}} w_i \K(v, \ve{r}|s_i,\sigma^s).
\end{equation}
further referred as the \emph{object model} as in \cite{kopicki2016}, where the weight $w_i=1/N_O$. Figure~\ref{fig:contact_models} shows in red the local area considered for extrapolating the features for two different contact types: a) flat contact and b) curved contact.

From the same point cloud, we collect a second set of $N_{T}$ features by uniformly sampling the visible point cloud to compute a second joint probability as 
\begin{equation} \label{eq:task_model}
    T(v, \ve{r}) \equiv P(v, \ve{r}) \simeq \sum_{j=1}^{N_{T}} w_j \K(v, \ve{r}|s_j,\sigma^t).
\end{equation}
further referred as the \emph{task model}, where $s_j=(v_j,\ve{r}_j)\in SE(3)\times\R^2$, $\sigma^t$ is the bandwidth and $w_j=r_{1j}/r$ is the weight associated with the feature represented by the $j$-th kernel and it is proportional to its highest curvature normalised with respect to the maximum curvature value available between the $T_s$ sampled points, $r$. This enables us to give more importance to salient features such as corners and edges.

\begin{figure}[t]
    \centering
    \includegraphics[width=.49\textwidth]{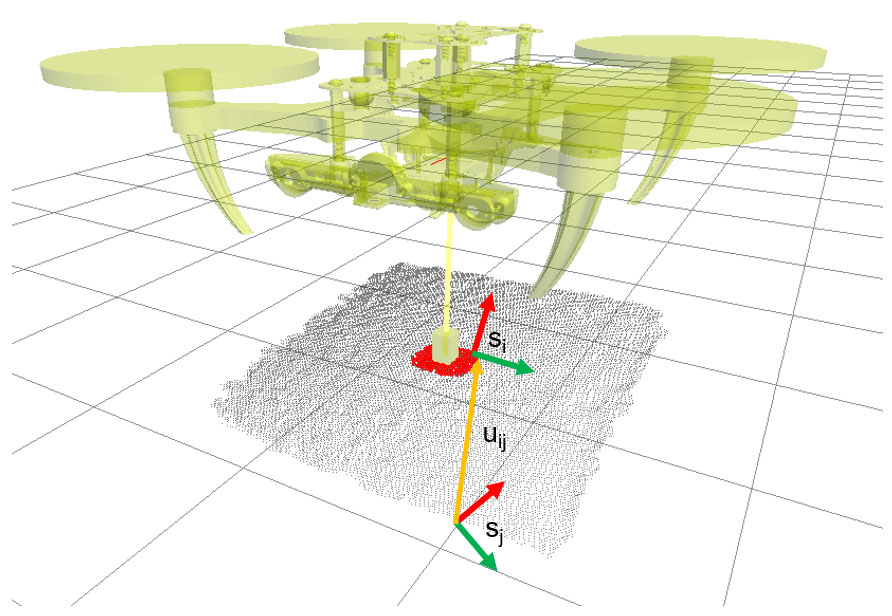}
    \includegraphics[width=.49\textwidth]{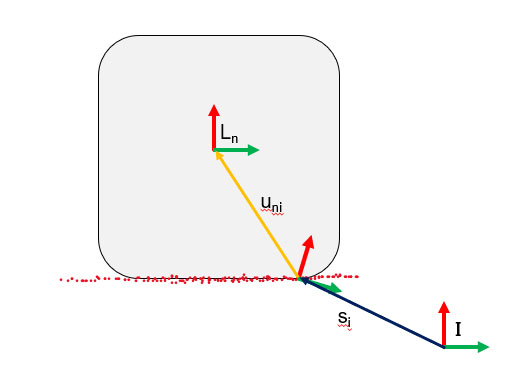}
    \caption{Graphical representation of the task model (left) and the link pose (right) used in the contact model. The task model is constructed as a rigid body transformation $u_{ij}$ in $SE(3)$ from a task feature $s_j$ to a surface feature $s_i$ in the contact region. The link pose is computed as a rigid body transformation $u_{ni}$ in $SE(3)$ from a  surface feature $s_i$ in the contact region to the reference frame of the robot's link. All the $SE(3)$ poses $s_j$, $s_i$, $L_n$ are computed with respect to the inertia frame $\mathcal{I}$.}
    \label{fig:task_model}
\end{figure}

\vspace{0.5cm}
\subsection{Contact models}\label{sec:contact_model}

The contact model describes the relation between the robot's end-effector and the object surface. It is trained in a one-shot fashion from a demonstration. A teacher presents to the robot the desired contact, either in simulation or in reality, then the local surface in contact with the robot's link is sampled and a probability density function is learned.

We denote by $u_{ni}=(\ve{p}_{ni},\ve{q}_{ni})\in SE(3)$ the pose $L_n$ of the $n$-th link relative to the pose $v_i$ of the $i$-th surface feature from \eqref{eq:object_model}. We compute this as
\begin{equation}
    u_{ni} = v_i^{-1}\circ L_n
\end{equation}
where $\circ$ denotes the pose composition operator, and $v_i^{-1}=(-q_i^{-1}p_i,q_i^{-1})$ is the inverse of pose $v_i$. 

The contact density $M_{n}(u, \ve{r})$ closely resembles the surface feature kernel function in \eqref{eq:surface_feature_kernel}, and is defined as follows:
\begin{equation}
    M_n(u, \ve{r}) \simeq \sum_{i=1}^{N_c} w_{ni} \mathcal{N}_3(\ve{p}\, | \, \ve{p}_{ni},\sigma_{p}^c)\ \Theta(\ve{q}\, |\, \ve{q}_{ni},\sigma_{q}^c)\ \mathcal{N}_2(\ve{r}\, |\, r_{ni},\sigma_{r}^c)
\end{equation}
where $N_c\leq N_O$ is a user defined parameter to allow downsampling to save computational time and $w_{ni}$ corresponds to the kernel's weighting such that $\sum_{i=1}^{N_c} w_{ni}\!=\!1$.

\vspace{0.5cm}
\subsection{Quadrotor configuration model}\label{sec:configuration}

The configuration model encodes the poses of the quadrotors and their grippers as demonstrated during training. For the case of a single drone, the configuration model merely describes the kinematic relation between the quadrotor and its gripper. However, when $n>1$ this model enables us to reduce the configuration space when transferring the contacts to another surface by focusing only on those configurations that resemble the one in the training example.

The poses of the $N_b\geq 1$ quadrotors are represented as $h_n=(b_n,L_n)$ with $b_n=(p_n,q_n)\in SE(3)$ the pose of the $n$-th drone and $L_n=(p_{L_n},q_{L_n})\in SE(3)$ the pose of its manipulative link.
We approximate the configuration density as
\begin{equation} \label{eq:contact_frame_probability}
    H(h) \simeq \sum_{i=1}^{N_b} w_i \mathcal{N}_3(\ve{p}_{L}\, | \, \ve{p}_{L_i},\sigma_{p})\ \Theta(\ve{q}_{L}\, |\, \ve{q}_{L_i},\sigma_{q}^c)\
\end{equation}
where $w_i=e^{-\alpha||b-b_i||^2}$ depends on the similarity between the drone's pose, $b$, and the kernel's one, $b_i$, while $L=(\ve{p}_L,\ve{q}_L)$ is the link pose compared against the $N_b$ link poses from the kernels, $L_i=(\ve{p}_{L_i},\ve{q}_{L_i})$.

\section{Transferring to novel surfaces}\label{sec:transferring}

Once the models are learned we can transfer the contacts on novel payloads. The aim is to find local features on the new object that are similar to the ones presented in the training and to place the robot manipulative links accordingly. For evaluating the approach, we learn several contact models for different tasks and contact types--we call them \emph{testing conditions} as introduced in \sect{sec:results}. However, when presented with a new point cloud, we manually select the appropriate contact model for each testing conditions. Automatic model selection can be formulated as an optimisation problem~\cite{optimisation} or a learning one~\cite{10.1177/0278364913495721}, but we kept it as out of scope for this work.

\begin{figure}[t]
    \centering
    \includegraphics[width=.99\textwidth]{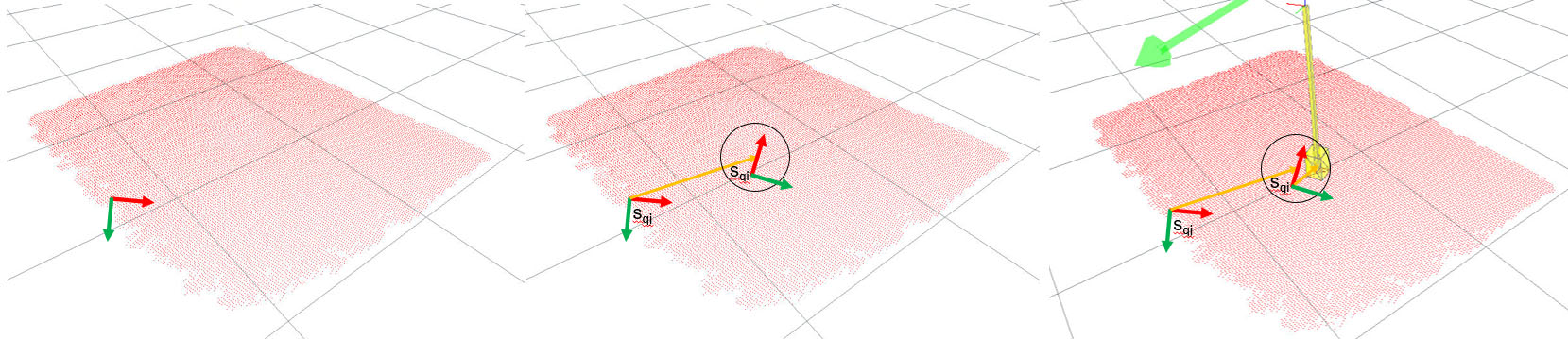}
    \caption{Graphical representation of inferring a candidate link pose for a novel surface. First, from the task model we sample a feature $s_{qj}$ on the novel point cloud. We then apply the transformation $u_{ij}$ observed when learning the model to identify a feasible contact region. From that region (circled area in the middle image), we sample a surface feature from the object model $O(v,\ve{r})$. Finally we apply the transformation $u_{ni}$ observed at training time to place the link $L_n$.}
    \label{fig:query_task_model}
\end{figure}

\vspace{0.5cm}
\subsection{Query density}\label{sec:query}

%\begin{figure}[t]
%    \centering
%    \includegraphics[width=.99\textwidth]{figures/one_drone_contact_results3.jpg}
%\caption{Inferring contacts on a new shape with a closed-to-edge contact. The first column shows the training example use to learn the contact model. The first and second rows show the transferring of the contacts for our approach. The other rows shows the modified baseline method from \cite{kopicki2016} which do not consider the task model from \sect{sec:object_model}. The first column shows the UAVs configuration, the surface used for learning and the contact features. Red points are the local features of the contacts, black points are the task features. The associated drone's pose is shown with a frame using the conventional RGB colour map for x-, y-, and z- axes, respectively.}
%\label{fig:dragging_contact_results}
%\end{figure}

The query density represents the distribution of link poses over a novel point cloud. By searching for local similarities in the query point cloud's surface features, we can estimate the relative link pose $u$ which transfers the learned contact onto the new surface. 
This process is designed to be transferable such that a model trained upon a single object can be applied to a variety of previously unseen objects. Figure \ref{fig:query_task_model} shows a graphical representation of the method: starting from a sampled task feature, we identify a potential area on the visible surface where to seek for a good contact.  

We define the query density as $Q(L_n,u,v,\ve{r})$ where $v\in SE(3)$ denotes a point on the objects surface expressed in the inertial frame, $\ve{r}\in\R^2$ is the surface curvature of such a point, $u\in SE(3)$ denotes the pose of the link relative to a local frame on the object, and $L_n$ is the pose of the $n$-th link with respect to the inertia frame.

For each link, its pose distribution over a new point cloud is given by marginalising $Q(L_n,u,v,\ve{r})$ with respect to $u$, $v$, and $\ve{r}$. Since $L_n=v\circ u$ is described in terms of $v$ and $u$, and by assuming that the density in the inertia frame of the surface point, $v$, and the distribution of link poses relative to a surface point, $u$, are conditionally independent given the curvature on a point, $\ve{r}$, we factorise the query density as follows
\begin{equation} \label{eq:query_density}
\begin{aligned}
    Q(L_n) = & \int\int\int P(L_n,u,v,\ve{r})dudvdr\\
    & \int\int\int P(L_n|u,v)P(u,\ve{r},v)dudvdr\\
    & \int\int\int P(L_n|u,v)P(u|\ve{r})P(v|\ve{r})P(\ve{r})dudvdr
\end{aligned}
\end{equation}
where, by following \cite{kopicki2016}, we implement $P(L_n|u,v)$ as a Dirac function, and $P(u|\ve{r})$ as $M_n(u|\ve{r})$, the conditional probability that the $n$-th link will be placed at pose $u$ with respect to the surface feature, given that this surface feature has curvature $\ve{r}$. $P(v|\ve{r})$ is implemented as $O(v|\ve{r})$, the probability of the observed curvature $\ve{r}$. Finally, $P(r)=M_n(\ve{r})O(\ve{r})$ is chosen to reinforce that the contact model $M$ and the object model $O$ are observing the same surface feature. $M_n(\ve{r})$ is the distribution of features in the contact model, and $O(\ve{r})$ is the distribution of feature $\ve{r}$ in the new point cloud.

\begin{figure}[t]
    \centering{
    \includegraphics[width=.99\textwidth]{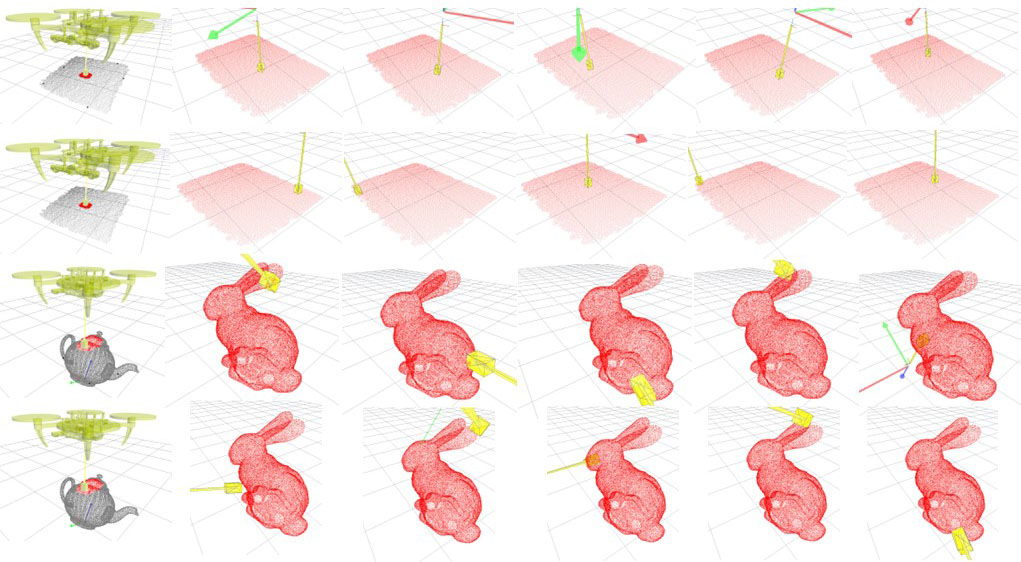}
    \includegraphics[width=.99\textwidth]{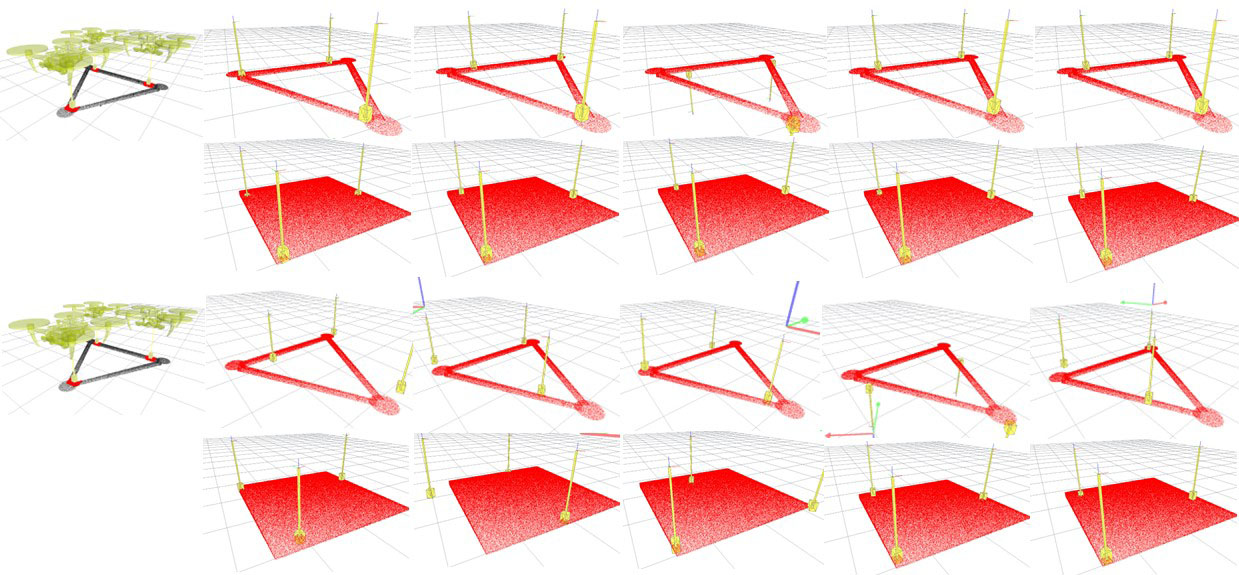}
    }
    \caption{Inferring contacts on a new shape for aerial manipulation. The first column shows the training examples use for learning the contact models. The first, third, fifth and sixth rows show the transferring of the contacts for our approach. The other rows shows the baseline method \cite{kopicki2016}. The first column shows the UAVs configuration, the surface used for learning and the contact features. Red points are the local features of the contacts, black points are the task features. The other columns demonstrate the link pose computed on a new surface. The associated drone's pose is shown with a frame using the conventional RGB colour map for x-, y-, and z- axes, respectively.}
    \label{fig:exp_contact_results}
\end{figure}

We extend this formulation with another random variable $u_{ij}=(\ve{v}_{ij}, \ve{q}_{ij})$ to encode the desired pose of the selected $i$-th surface feature of the contact model relative to the task model's features defined in \eqref{eq:task_model}. Thus, $u_{ij} = v_i^{-1}\circ v_j$, where $s_j=(v_j,q_j)$ belong to the task model.
Then we approximate the query density by sampling $N_Q$ kernels centred on weighted link poses from \eqref{eq:contact_frame_probability}, so that   
\begin{equation} \label{eq:query}
    Q(L_n) \simeq \sum_{k=1}^{N_q} w_{nk} \mathcal{N}_3(\ve{p}\, | \, \ve{p}_k,\sigma_{p})\ \Theta(\ve{q}\, |\, \ve{q}_k,\sigma_{q})
\end{equation}
where $(\ve{p}_k,\ve{q}_k)\in SE(3)$ describes the $k$-th kernel. The variable 
$$
w_{nk}=\frac{1}{Z}\sum_{i=1}^{N_i}P(L_n|u_{ni},v_i)M(u_{ni}|\ve{r}_i)M(\ve{r}_i)O(v_i|\ve{r}_i)O(\ve{r}_i)(\sum_{j=1}^{N_j}P(u_{ij}|\ve{r}_i,\ve{r}_j)T(v_j|\ve{r}_j)T(\ve{r}_j))
$$
weights the query density according to the contact model and the task model. To compute the weighting, we randomly sample from the new point cloud $N_i$ surface features from the estimated contact surface and $N_j$ features according to the task model so that $s_j\sim T(v_j|\ve{r}_j)$ is the observed feature $\ve{r}_j$ on the new point cloud. Again, we implement $P(u_{ij}|\ve{r}_i,\ve{r}_j)$ as a Dirac function and $T(v_j|\ve{r}_j)T(\ve{r}_j)$ represents the probability of the observed curvature on the new point cloud. The value $Z$ is a normaliser, and the value $N_i$ and $N_j$ are maintained constant across all the experiments.

\vspace{0.5cm}
\subsection{Contact optimisation and selection}\label{sec:opt}

Let us denote by $\hat{h}=\{(b_i,L_i)\}_i^n$ the poses of the quadrotors and their gripper in $SE(3)$. The objective of the contact optimisation is to generate candidate contacts by maximising the product of the query densities and the quadrotors configuration density, as follows:
\begin{equation} \label{eq:product}
    \argmax_{\hat{b}}\Jcal(\hat{h})=\argmax_{(b_i,L_i)} \prod_{(b_i,L_i)\in\hat{b}} H(h_i)Q(L_i)
\end{equation}
where $\Jcal$ is the likelihood of the candidate contact on the new point cloud. We optimise the likelihood using simulated annealing (SA) \cite{sa}.

\section{Results}\label{sec:results}

\begin{figure}[t]
\centering
\includegraphics[width=.98\textwidth]{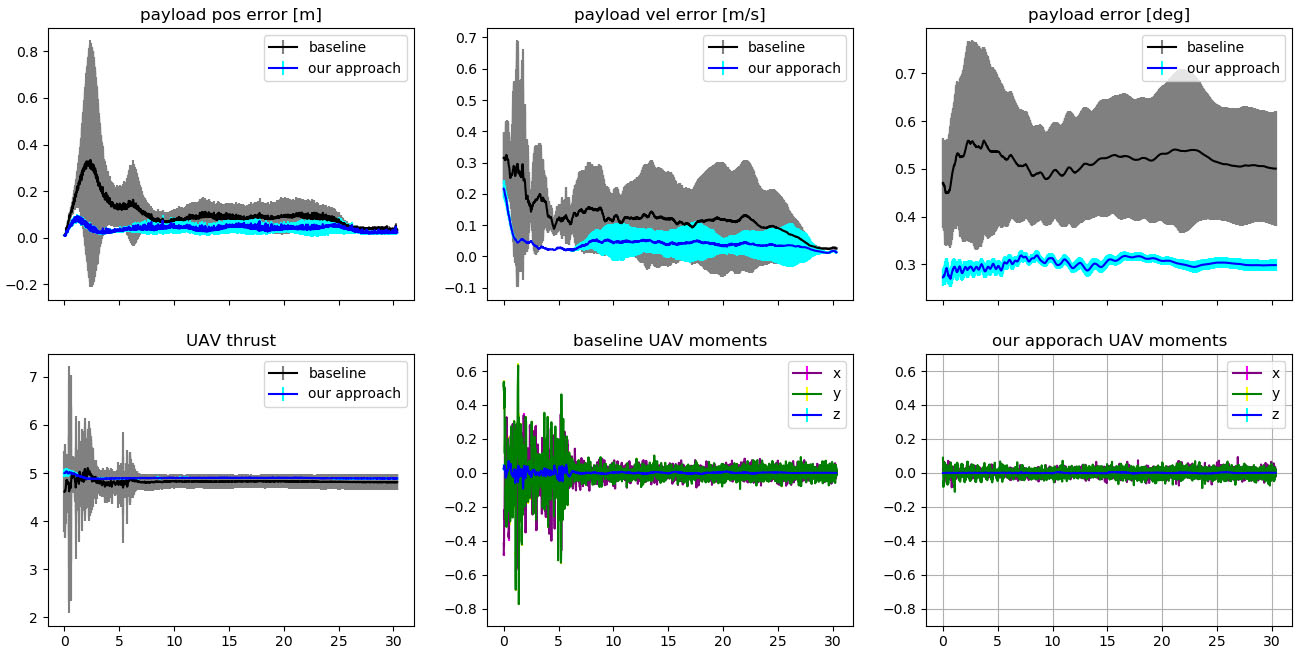}
\caption{The results summarise the transportation experiments. Each solution from \fig{fig:exp_contact_results} are used for 10 trials in a simulated environment in which the single or multiple UAVs lift the payload, transport it along a circular trajectory twice, and then hover the payload above ground for at least five seconds. The x-axis represent the time in second for all plots. The top row shows the averaged position and velocity error w.r.t. the desired payload's trajectory (baseline and proposed method separately), and both averaged rotational errors together in the last plot. The bottom row show the UAV's thrust (together) and moments (separately). For the three UAVs condition, the values in the bottom row have been averaged for each UAV.}
\label{fig:sim_results}
\end{figure}

\begin{table}[t]
\centering
\caption{Experimental parameters. All of the parameters are kept constant over the experiments. We use diagonal inertia matrices for both the UAVs and payloads. The gains values are for the $x-$, $y-$ and $z-$ components respectively. UAV and gains parameters are taken from \cite{loianno_pcmpc_2021}.}
\label{tab:parameters}
{\renewcommand{\arraystretch}{1.2} %<- modify value to suit your needs
\begin{tabular}{ ||c | c ||c | c ||c | c ||c | c | c||}
\hline
\hline
\multicolumn{2}{||c||}{Contacts} & \multicolumn{2}{c||}{UAV} & \multicolumn{2}{c||}{Payload} & \multicolumn{3}{c||}{Gains} \\
\hline
\hline
\multicolumn{2}{||c||}{Learning} & $m_n$ & 250g & $m_z$ & 250g & & Controller & Cable \\
\hline
$N_O$ & 500 & $\Jx_n^{xx}$ & 0.0006 & $\Jx_z^{xx}$ & 0.006 & pos & 6, 6, 10 & 1.45, 1.45, 3 \\
\hline
$N_T$ & 50 & $\Jx_n^{yy}$ & 0.0005 & $\Jx_z^{yy}$ & 0.0059 & vel & 3, 3, 6 & 3.7, 3.2, 2.5 \\
\hline
$N_C$ & 500 & $\Jx_n^{zz}$ & 0.001 & $\Jx_z^{zz}$ & 0.107 & rot & 0.25, 0.25, 0.08  & 0.5, 0.5, 0.02\\
\hline
\multicolumn{2}{||c||}{Inferring}  & $l_n$  & 0.5m  & \multicolumn{2}{c||}{} & ang & 0.025, 0.025, 0.005 & 0.16, 0.16, 0.04 \\
\hline
$N_i$ & 500 & \# prop & 4 &  \multicolumn{2}{c||}{}   & $\xi$ & 16, 16, 16  & \\
\hline
$N_j$ & 5 & max RPM & 16400 &  \multicolumn{2}{c||}{}  & $\omega$ & 6.7, 6.7, 3.1 & \\
\hline
$N_Q$ & 1000 & min RPM & 5500 &   \multicolumn{2}{c||}{}   &  \multicolumn{3}{c||}{}  \\
\hline
\hline
\end{tabular}
}
\end{table}

The generalisation to novel objects of the contact models has been evaluated in a set of simulated experiments on several conditions as follows. For a single UAV, we learn two types of contacts: (i) a contact placed at the centre of the visible upper surface of a box, and (ii) a contact on a curved surface.
For multiple drones, we learn a triangular formation for three UAVs for a transportation task.

In all conditions, we learn each contact model in one single demonstration. Each training object was loaded in the simulation as full or partial point cloud, as presented in~\fig{fig:contact_models}. We do not retrieve global information about the object, such as its CoM. The UAVs were manually placed to generate the desired contacts with their passive gripper over the visible point cloud. For all conditions, we sampled $N_O=500$ local features to represent the contacts, in both our approach and baseline method, and $N_T=50$ for the task model (present only in our approach). Once a contact model was learned, it was stored and labelled for future reference. Only kinematic and geometrical information were used to learn the models; forces and dynamics are not needed for learning.

Once the contact models are learned, a new full or partial point cloud is presented to the system. In~\fig{fig:exp_contact_results} we show the point clouds belonging to four objects used for testing: a) 68cm x 61 cm x 28cm FedEx box (top two rows); b) the Stanford bunny (third and fourth rows); the hollowed triangular shape used to learn the three drones' contact models (fifth and seventh rows); and a 68cm x 68 cm x 28cm FedEx box (sixth and eighth rows). We then compare the best five solutions generated for each condition with the ones computed by an adapted version of the grasping algorithm presented in \cite{kopicki2016}. Since \cite{kopicki2016} considers a robot manipulator equipped with a hand, we modify it to consider the quadrotor configuration model. Therefore, the main difference between the two approaches remains the task model.

Furthermore, each solution from \fig{fig:exp_contact_results} is also evaluated in a simulated environment for a transportation task. The simulator is initialised with the computed optimised contacts, and the UAVs will need to transport the payload along a pre-planned circular trajectory with a 1m radius performed for twelve seconds. The trajectory is computed such that the UAVs need to lift the payload to the starting position of the trajectory, perform the circle twice and then hover above the ground, maintaining the payload in the same position as the last waypoint of the trajectory. The simulation uses the PCMPC controller as in \cite{loianno_pcmpc_2021}, also adapted by us to control a single UAV with a payload that is not a point mass object. Additional, Gaussian noise with 1cm standard deviation was applied to the perceived position of the payload, and each transportation task for each contact was repeated ten times. The results were averaged across each object and trial and presented in \fig{fig:sim_results}. 

The solutions for the hollowed triangular shape are from the complete point cloud of the object, for which solution 3 for our approach and solution 4 for the baseline suggested an upside-down contact. We discarded these solutions from the transportation task since we could not achieve the required kinematic of the contact. All the other solutions from the baseline methods that do not provide contacts for each UAV, have been modified to connect the gripper to the closest point on the payload.
Considering the kinematically infeasible solutions for the three drones condition, we had a total of 147 runs (3 contact models x 5 solutions x 10 trials $-$ 3 infeasible solutions) for the proposed model and 147 for the baseline. In all conditions we use a set of fixed parameters as shown in \tab{tab:parameters}. The next section will discuss our results and conclude with our final remarks.

\section{Conclusion}\label{sec:conclusion}

This paper proposed a framework for learning task-dependent contacts for aerial manipulation. Our models are learned in a one-shot fashion and do not require a complete CAD model of the payload or dynamic modelling. We build on a contact-based formulation. Typically, such methods rely only on local information, and task-dependent features must be handcrafted. In contrast, thanks to the task model presented in \sect{sec:object_model}, we capture meta-information regarding the task by merely looking at the geometrical features of the point cloud, without the need for user-specific insight about the task. 

In \fig{fig:exp_contact_results} the second row shows the solutions from \cite{kopicki2016}, which tend to prefer contacts in feature-rich areas of the new query (i.e. the edges), while our approach forces the contacts to lay in the middle; ideal for transporting a box with a uniform mass distribution by a single drone. More importantly, it is consistent with the demonstrated example by the expert at training time; a flat contact in the middle of the visible surface. The same principle is visible in the third and fourth rows, where the contacts are transferred from a teapot to a more complex shape: the Stanford bunny. Again, our approach tends to lay the contacts as close as possible to the middle section of the bunny, which minimises the payload's oscillations during the experiments. The CoM for the bunny is computed as the reference frame of the STL model, which lays on the neck of the bunny, making contacts on its head the most effective for transportation, and contacts along its middle section the second most effective. This information is not available to the learned models. Therefore, the first solution generated by the baseline on the bunny's thigh is the worse type of contact, and the UAV was incapable of controlling the payload configuration accurately, which rotated on itself and swung along the entire trajectory. 

In the case of the three UAVs, the search space is more constrained by the demonstrated configuration of the UAVs. The baseline, driven by the only aim of maximising the similarity with respect to the observed contacts nearby the gripper at training time, often gets stuck in local minima providing sub-optimal solutions, in which one or two good contacts in feature-rich areas lead to higher likelihoods than three reasonable ones. Our approach tends to discard those local minima in the tested examples thanks to the task model by constraining the choice of the selected contact points. 

In \fig{fig:sim_results} we observe the averaged difference in transporting the payloads given the choice of the contacts. Our approach outperforms the baseline in terms of accuracy in following the pre-planned payload trajectory and by minimising the effort computed by the UAVs as thrust and moments. It is interesting to observe that, in all conditions, our approach enables the UAV to control the orientation of the payload in a more agile way. This is due to the relative position of the contact points with respect to the payload's shape, which minimises oscillations during the transportation. This is clearly visible by the lower standard deviation from the desired position and orientation of the payload associated with our approach. Furthermore, the velocity profile of the payload and the output to the controller (thrust and moments) show that our contact points simplify the work of the controller in following the desired path.  

Although performed in simulation, the empirical evaluation shows the potential of the proposed idea. Without physical and dynamic information about the system, a full convergence to an optimal solution is impossible to achieve. Nonetheless, such information is rarely available in reality, and this work is the first step towards autonomous contact selection for aerial manipulation. 

\section{Strengths, limitations and future work}

The presented work extends a (static) contact-based formulation of grasp synthesis to a dynamic task. While previous efforts have focused on integrating an approximation of such dynamics in the models, we investigate how much knowledge about the task we can capture by disregarding the dynamics altogether. 

The probabilistic formulation could be considered as a soft simultaneous optimisation of multiple experts. Each expert learns some of the characteristics of the contacts within the feature space, e.g. the task model, and weighs their opinion in the choice of the candidate contact at query time. Any soft simultaneous optimisation of multiple criteria would work in principle. However, the probabilistic representation allows us to draw contact points from a continuous representation of the payload's surface. Nevertheless, the experts only weight geometrical properties, which do not play a key role in defining the dynamic behaviour of the payload. We assume that the demonstration encodes the contact type (e.g., flat contact), the task (e.g., contact above the payload's CoM for a transportation task) and the visible surface of the object provides enough salient features to learn a good task model. Although there is no need for handcrafting task features into the model, the choice of the training data becomes, therefore, crucial.

Furthermore, our approach assumes that the contact types and tasks are conditionally independent in our model. However, this is generally not true, as seen in the transportation of a FedEx box and the teapot examples. Applying the model learned on the FedEx over the Stanford bunny will lead to a poor choice of contacts since flat surfaces are quite limited in the chosen payload. Therefore, generalisation across contact types and tasks becomes harder to achieve.
This also limits the use of the same contact model for a different task (e.g., flat contact next to the edge for dragging the payload) which will require learning a new model. 

With many models that need to be learned, we also need to solve the problem of model selection when facing a novel payload. In our empirical evaluation, we disregard this problem by providing the correct model manually at testing time. Reinforcement learning techniques can be used for model selection, but the task would need to be encoded in a more informative manner for this technique to converge to a good reward function.     

The choice of the principal curvatures as surface features is also limiting. It is a baseline for determining the contact points and the general shape of the payload, but other features may provide better information about the intrinsic dynamic properties of the payload or the encoded task, such as surface roughness or intensity features--or more likely a combination of multiple features.

In order to achieve a fully autonomous aerial manipulation, all these problems will need to be addressed. This work pioneers the problem of grasp synthesis for aerial transportation, but with the increasing interest in aerial manipulation and the technological maturity that we are witnessing, we would suggest that it will become central to many applications. In future work, we will focus on evaluating this framework on a real platform with payloads with different dynamical properties, e.g., mass distribution. Since our approach is a generative model, multiple candidate contacts will be provided for each payload. Primitive dynamic motions could be used to evaluate the selected contacts according to the observed behaviour of the payload.

%%%%%%%%%%%%%%%%%%%%%%%%%%%%%%%%%%%%%%%%%%%%%%%%%%%%%%%%%%%%

%Even more challenging problem is to plan multi-contact manipulations. In this case, the interaction of different forces which are applied simultaneously to an object has to be modelled. It is possible to tune a physics engine to produce realistic simulation.

\section*{Author Contributions}

%The Author Contributions section is mandatory for all articles, including articles by sole authors. If an appropriate statement is not provided on submission, a standard one will be inserted during the production process. The Author Contributions statement must describe the contributions of individual authors referred to by their initials and, in doing so, all authors agree to be accountable for the content of the work. Please see  \href{http://home.frontiersin.org/about/author-guidelines#AuthorandContributors}{here} for full authorship criteria.
CZ is the main author of this work. CZ has designed and implemented the idea, evaluated the framework and written the manuscript. EF has supervised this project and revised the manuscript and data. 
%RH has collaborated in finding the literature and final editing. RS has funded this project.

%\section*{Funding}
%This work was supported by UK Engineering and Physical Sciences Research Council (EPSRC No. EP/R02572X/1) for the National Centre for Nuclear Robotics (NCNR).
%Details of all funding sources should be provided, including grant numbers if applicable. Please ensure to add all necessary funding information, as after publication this is no longer possible.

\section*{Acknowledgments}
We thank Giuseppe Loianno and Marek Kopicki for the support in developing this work.
%This is a short text to acknowledge the contributions of specific colleagues, institutions, or agencies that aided the efforts of the authors.

%\section*{Supplemental Data}
% \href{http://home.frontiersin.org/about/author-guidelines#SupplementaryMaterial}{Supplementary Material} should be uploaded separately on submission, if there are Supplementary Figures, please include the caption in the same file as the figure. LaTeX Supplementary Material templates can be found in the Frontiers LaTeX folder.

%\section*{Data Availability Statement}
%The datasets [GENERATED/ANALYZED] for this study can be found in the [NAME OF REPOSITORY] [LINK].
% Please see the availability of data guidelines for more information, at https://www.frontiersin.org/about/author-guidelines#AvailabilityofData

\bibliographystyle{plain} % for Science, Engineering and Humanities and Social Sciences articles, for Humanities and Social Sciences articles please include page numbers in the in-text citations
\bibliography{references}

\end{document}